\documentclass[11pt,a4paper]{article}
\usepackage{times,latexsym}
\usepackage{tabularx}
\usepackage{graphicx}
\usepackage{algorithm}
\usepackage{algorithmic}
\usepackage{tikz}
\usepackage{xcolor}

\usepackage{fontspec}
\usepackage{xeCJK}
\usepackage[acceptedWithA]{tacl2021v1}
\usepackage{xspace,mfirstuc,tabulary}

\newif\iftaclinstructions
\taclinstructionsfalse % AUTHORS: do NOT set this to true
\iftaclinstructions
\renewcommand{\confidential}{}
\renewcommand{\anonsubtext}{(No author info supplied here, for consistency with
TACL-submission anonymization requirements)}
\newcommand{\instr}
\fi

\iftaclpubformat % this "if" is set by the choice of options

\else

\fi

\title{From Inline Notes to Collected Commentaries: Toward Context-Preserving Organization of Exegetical Knowledge in Classical Chinese Texts}

\author{
Ke Liang$^{1}$, Qi Su$^{2, 3, 4}$, Churen Huang$^{1}$
\\[0.5ex]
$^{1}$Department of Language Science and Technology, The Hong Kong Polytechnic University
\\[0.5ex]
$^{2,3,4}$School of Foreign Languages; Institute for Artificial Intelligence; \\ Research Center for Digital Humanities, Peking University
\\[0.5ex]
\texttt{\{leo-ke.liang,churen.huang\}@connect.polyu.hk}
\\
\texttt{sukia@pku.edu.cn}
}

\date{}

\begin{document}
\maketitle
\begin{abstract}
  Inline notes and collected commentaries are important forms of scholarly communication that evolved within the Confucian exegetical tradition, yet have received little computational attention. Drawing on traditional Chinese exegetics and philology, this paper formulates collected commentary compilation as an NLP task and proposes a computational framework that preserves the contextual dependency of inline notes while enabling their automatic compilation and exegetical knowledge organization. It combines two-step prompt chaining for identifying the associated main-text segments and exegetical functions of annotations with cross-source mention clustering for integrating commentary across editions, achieving a CoNLL F$_1$ score above 97\% in a case study on the \textit{Classic of Mountains}. Our framework lays the foundation for the large-scale organization of historical exegetical knowledge, thereby supporting a broad range of downstream philological and NLP tasks.
\end{abstract}

% Submission-specific rules
\section{Introduction}
Commentaries on classical texts are explanatory texts that developed alongside the transmission of these works, aiming to interpret their linguistic \citep{LingSystematicXungu2021} and historical meanings \citep{DongZhuShiYuXunGu1993a}. They not only facilitate the understand of ancient culture, but also play an important role in natural language processing (NLP) for ancient text, yet this potential remains largely underexplored. In this paper, we explore the computational organization of ancient commentaries based on their intrinsic characteristics. The resulting resources could be used for both philological downstream tasks and NLP applications.

The tradition of annotating classical texts can be traced back to the Spring and Autumn period (770-476 BCE). Owing to the Confucian emphasis on reverence for antiquity, it became the dominant mode of scholarly writing from the Western Han dynasty (206 BCE-9 CE) throughout the premodern period \citep{gardnerConfucianCommentaryChinese1998}. Unlike the Western tradition of marginalia or footnotes, Chinese classical commentaries have, since the Han Dynasty, been embedded within the main text as inline notes (隨文注釋) and differentiated from it by smaller double-column script \citep{wilkinsonChineseHistoryNew2018}. As commentaries accumulated, scholars often needed to consult multiple commentarial editions (注本), each containing a distinct body of commentary, to obtain a comprehensive understanding of a classical text, a process that was both complicated and time-consuming. To address this challenge, scholars in the Eastern Han (25-220 CE) began integrating commentaries across different editions into a single base edition, sometimes supplemented by the compiler's own interpretations \citep{ZhangDefinitionCollectedAnnotations2011}, giving rise to the tradition of collected commentaries, or \textit{jijie} (集解, also \textit{jizhu} 集注, or \textit{jishi} 集釋), which was subsequently adopted to numerous Confucian and non-Confucian classics \citep{YingStudyShijiJijie2005}. Having last for nearly 2000 years, this tradition remains valuable for organizing the vast body of historical commentaries in today's digital age.

However, automatically compiling collected commentaries is far from straightforward due to the contextual exegesis (隨文釋義) practiced in inline notes, which means that inline notes can only be interpreted within their original context. Successful compilation needs to preserve their contextual dependency after the detachment and reattachment, requiring consideration of the main-text variants, commentary placement, and potential repetition or contradiction across editions. Failure to do so may lead to inconsistency and imprecision, as observed in the \textit{Shiji} \citep{MaReexaminationOriginTwoAnnotation2022}.

In this paper, we formulate the compilation of collected commentary as an NLP task and propose a context-preserving computational framework for its automatic compilation and exegetical knowledge organization. Drawing on Chinese exegetics and philology, we derive a set of criteria for collected commentary compilation, among which identifying the main-text segment associated with each commentary is particularly challenging. Accordingly, we decompose the framework into two subtasks: (1) large language model (LLM)-based identification of associated main-text segments and exegetical functions; and (2) semantic similarity-based cross-source mention clustering for commentaries from different editions. To implement this framework, we take the \textit{Classic of Mountains} (山經, \textit{Shan Jing}, hereafter \textit{SJ}), one section of the \textit{Classic of Mountains and Seas} (山海經, \textit{Shanhai Jing}, hereafter \textit{SHJ}) \footnote{We focus on the \textit{SJ} because of its intrinsic differences from the other parts of the \textit{SHJ} \citep{ChenShanHaiJingXueShuShiKaoLun2012} and downstream-task considerations.}, as a case study. Our framework achieves a CoNLL F$_1$ score above 97\%.

Our emphasis on contextual dependency reflects not only the intrinsic nature of traditional commentaries but also the context-sensitive nature of language, aligning with the development of contextual representation in NLP \citep{vaswaniAttentionAllYou2017, petersDeepContextualizedWord2018, devlinBERTPretrainingDeep2019, choiDecontextualizationMakingSentences2021}. The main contributions of this paper are twofold. First, grounded in Chinese exegetics and philology, we formulate collected commentary compilation as an NLP task and propose a computational framework for its compilation and exegetical knowledge organization. Second, our framework provides a general approach to organizing large-scale historical exegetical knowledge from collected commentaries. The resulting structured knowledge can serve as a foundation for a broad range of downstream philological and NLP tasks, facilitating both humanities research on classical Chinese and the adaptation of modern NLP methods to classical texts.

\section{Related Work}
Many NLP tasks have been applied to ancient texts, whereas ancient commentaries have received comparatively little attention. In this section, we review previous work in two areas—NLP research on ancient classics and the computational processing of ancient commentaries, with a particular focus on Chinese—and position our work within this literature.

\subsection{NLP Research on Ancient Classics}
Regarding NLP research on ancient Chinese classics, early studies mainly focused on tasks such as part-of-speech tagging (POS), Chinese word segmentation (CWS), named entity recognition (NER) \citep{huangClassicalChineseSentence2010, tangResearchAutomaticallyRecognizing2013, huangResearchConstructingAutomatic2015, wangSentenceSegmentationMethod2016, huangPragmaticApproachClassical2018, liCapsulesBasedChinese2018}, mainly based on rule-based approaches or traditional machine learning methods. More recently, Transformer architecture and subsequent LMs have substantially advanced research on these tasks \citep{huKnowledgeRepresentationSentence2021, tangThatSlepenNyght2022,WangConstructionApplicationPretrained2022, wangUncertaintybasedRetrievalFramework2022,yasuokaSequenceLabelingRoBERTaModel2023a, zhuResearchWordSegmentation,tangCHisIECInformationExtraction2024,xuSemanticenhancedGraphNeural2024a,dengDiachronicNamedEntity2024,kangResearchNamedEntity2025,tangLanguageModelCollaboration2026}. For example, \citet{huKnowledgeRepresentationSentence2021} employed BERT-CRF and BERT-CNN models for CWS on ancient texts. \citet{tangCHisIECInformationExtraction2024} constructed a cross-era dataset for developing and evaluating NER and relation extraction. \citet{dengDiachronicNamedEntity2024} integrated temporal features into PLMs for named entity disambiguation in historical documents. 

Besides these task-oriented studies, considerable effort has also been devoted to constructing specialized linguistic resources, including a word sense annotation corpus \citep{shuConstructionApplicationAncient2022}, ancient Chinese allusion resources \citep{moConstructionApplicationAncient2024}, and a Tongjiazi (通假字) dataset \citep{wangAncientChineseLanguage2024}. These resources provide valuable training datasets for downstream NLP models. Additionally, recent studies have further explored knowledge-oriented applications, such as knowledge graph \citep{liangTextHistoricalEcological2024}, GraphRAG \citep{yangResearchGraphretrievalAugmented2026}, multimodal fusion classification model \citep{shenNewCompendiumMyriad2026}, aiming to facilitate knowledge retrieval, reasoning, and intelligent question answering over classical texts.

\subsection{Commentary Processing}
Regarding ancient commentaries processing, to our knowledge, computational studies of Chinese commentaries began with \citet{MaAutomaticAnalysisComments2012, MaConstructionAnalectsConfucius2012}, who performed automatic analysis of commentaries on the \textit{Analects}. Their workflow consisted of sentence alignment, which aligns the main text across different editions at sentence level, and annotation alignment, which identifies the main-text segment associated with each annotation \citep{ChenXianQinWenXian2013}. Similar approaches were subsequently applied to other Confucian classics, including the \textit{Yijing} \citep{JiaZhiShiLeiJu2015}, \textit{Shijing} \citep{WangStudySentenceAlignment2018}, \textit{Zuozhuan} \citep{XuStudyAutomaticAlignment2019}, and \textit{Mengzi} \citep{LiangZhuShuMengZi2021}. More recently, \citet{liStructureawareApproach2025} formulated annotation alignment as a semantic graph construction problem, using modularity-maximization clustering to align annotations with their corresponding main-text sentences within each individual edition.

Comparable concerns also appear in Western digital humanities research on medieval fragments \citep{bertiLinkedFragmentTEI2014/Dec/28, bertiLeipzigOpenFragmentary2016, dobchevaManuscriptFragmentsUniversity2018}, biblical catenae \citep{paparnakisDigitalGreekPatristic2017}, the Commentaria in \textit{Aristotelem Graeca et Byzantina} \citep{CAGBDigitalHumanities}, and the scholia of the \textit{Venetus A} manuscript \citep{HomerMultitextProject}. Although these studies primarily focus digitization, they likewise highlight the central role of annotations in preserving and interpreting ancient knowledge.

\subsection{Research Positioning}
Existing studies on commentary processing have established several core research tasks, including automatic commentary analysis and annotation alignment, and have identified challenges such as main-text variants \citep{XuStudyAutomaticAlignment2019} and weak semantic similarity between targets and explanations \citep{liStructureawareApproach2025}. However, little attention has been paid to collected commentaries, the contextual dependency of inline notes, and the context-preserving organization of exegetical knowledge.

More importantly, most NLP studies reviewed above, from NER to knowledge graph construction, rely on manually curated datasets \citep{huangClassicalChineseSentence2010, huangResearchConstructingAutomatic2015, liCapsulesBasedChinese2018, tangThatSlepenNyght2022, zhuResearchWordSegmentation,dengDiachronicNamedEntity2024, kangResearchNamedEntity2025,liangTextHistoricalEcological2024}, whose construction often requires expert knowledge derived directly or indirectly from commentaries. Research has also demonstrated that external knowledge \citep{nieBorrowingWisdomWorld2022}, including commentarial information \citep{XuSegmentationZuo2012} can facilitate NER and CWS. Therefore, our work on collected commentary compilation and exegetical knowledge organization may provide a foundation for future development in these tasks.

\section{Theoretical Foundations}
This section outlines the theoretical foundations for the computational organization of historical exegetical knowledge, covering commentary hierarchy and placement, the contextual dependency of inline notes, exegetical functions, and integration criteria. Because Chinese commentary systems vary across texts and require individual analysis \citep{NiXiaoKanXue2004}, we focuse on the \textit{SJ} while drawing on other classical texts to illustrate the broader applicability.

\begin{figure}[t]
    \centering
    \includegraphics[width=\columnwidth]{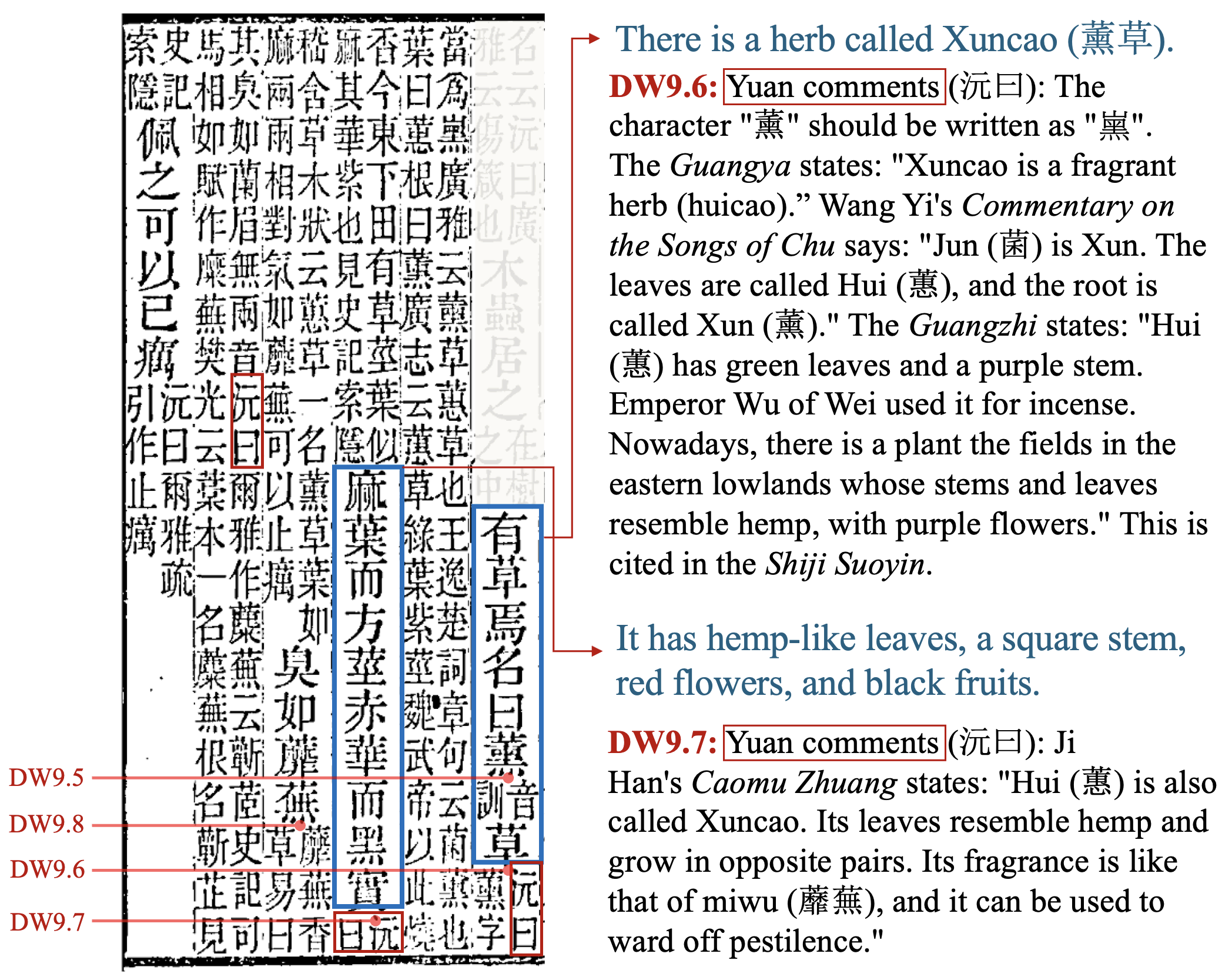}
    \caption{The \emph{Xuncao} passage of Edition D.}
    \label{fig:xuncao}
\end{figure}

Table~\ref{tab:SHJ-edition} lists the five \textit{SHJ} editions examined in this study (Editions A-E): Guo Pu's, Wu Renchen's, Wang Fu's, Bi Yuan's and Hao Yixing's commentaries (hereafter Guo, Wu, etc.). In all five editions, the commentaries are embedded within the main text in smaller double-column script. Figure~\ref{fig:xuncao} shows a page excerpt on \emph{xuncao} (薰草) from Edition D and part of its corresponding translation. Table~\ref{tab:qiuru_excerpt} presents the digitized main text and selected commentaries from the “Qiuru Mountain” paragraph across the five editions. Character variants in the main text are marked with red circles. Each commentary is assigned a unique ID. These examples are used throughout to illustrate the theoretical foundations.

\begin{table*}[!t]
%\begin{center}
\centering
%\small
\scriptsize
\begin{tabularx}{\textwidth}{%
>{\raggedright\arraybackslash}p{1.1cm}
>{\raggedright\arraybackslash}p{1.3cm}
>{\raggedright\arraybackslash}p{2.4cm}
>{\raggedright\arraybackslash}p{1.0cm}
>{\raggedright\arraybackslash}X
>{\raggedright\arraybackslash}p{0.2cm}
}
\hline
\multicolumn{1}{c}{\textbf{Dynasty}} &
\multicolumn{1}{c}{\textbf{Annotator}} &
\multicolumn{1}{c}{\textbf{Name}} &
\multicolumn{1}{c}{\textbf{Dates}} &
\multicolumn{1}{c}{\textbf{Edition}} &
\multicolumn{1}{c}{\textbf{Tag}} \\
\hline
Jin 晋 & Guo Pu 郭璞 & \textit{Annotations on the SHJ} 山海經傳 & $\ge$ 321 & \textit{Sibu Congkan} (First Series), reproduction of a Ming Chenghua gengyin (1470) edition from the Shuangjianlou (雙鑑樓) collection of Fu of Jiangan. & A \\
Qing 清 & Wu Renchen 吳任臣 & \textit{Extensive Annotations on the SHJ} ～廣注 & 1667 & Jin-chang Shuye Tang (金閶書業堂) edition, engraved in Qianlong 51 (1786); copy held at the Harvard-Yenching Library. & B \\
Qing 清 & Wang Fu 汪紱 & \textit{Preserved Texts of the SHJ} ～存 & $<$ 1759 & Lithographic edition issued by Lixue Zhai (立雪齋), Guangxu 21 (1895). & C \\
Qing 清 & Bi Yuan 畢沅 & \textit{Newly Revision of the SHJ} ～新校正 & 1781 & Zhejiang Shuju edition, Guangxu 3 (1877), based on the Lingyanshan Guan (靈巖山館) copy of the Bi. & D \\
Qing 清 & Hao Yixing 郝懿行 & \textit{Annotation and Commentary on the SHJ} ～箋疏 & 1804 & Supplementary re-engraved edition of \textit{Haoshi Yishu} (郝氏遺書), Guangxu 7 (1881); copy held at the Tsinghua University Library. & E \\
\hline
\end{tabularx}
%\end{center}
\caption{\label{tab:SHJ-edition} Selected inline-commentary versions on the \textit{SHJ}}
\end{table*}

%\begin{figure*}[!t]
%    \centering
%    \includegraphics[width=\textwidth]{figures/Fig1-printing-edition.png}
%    \caption{The “Qiuru Mountain” paragraph in the five annotated versions.}
%    \label{fig:printing-edition}
%\end{figure*}

\newcommand{\circchar}[1]{%
\tikz[baseline=(C.base), scale=0.78, transform shape]{
\node[draw=red!70, circle, line width=0.4, inner sep=0.1pt](C){\strut #1};
}
\hspace{-0.3em}%
}

\newcommand{\boxlabel}[1]{%
\tikz[baseline=(B.base), scale=0.78, transform shape]{
\node[draw=red!70, rectangle, rounded corners=0.2pt, line width=0.4pt, inner xsep=0.15pt, inner ysep=0.15pt](B){\strut #1};
}
\hspace{-0.3em}%
}

\newcommand{\annote}[1]{%
  \textsuperscript{\textcolor{blue!70!black}{[#1]}}%
  \hspace{-0.3em}%
}

\newcommand{\trans}[1]{%
{\fontsize{7pt}{10pt}\selectfont #1}%
}

\begin{table*}[!t]
\centering
%\footnotesize
\scriptsize
%\begin{tabular}{p{0.04\textwidth} p{0.90\textwidth}}
\begin{tabularx}{\textwidth}{@{}X@{}}
\hline
\multicolumn{1}{c}{\textbf{Main Text and Commentaries}}\\
\hline
\colorbox{gray!20}{A-Text:} 又北二百五十里，曰求如之山，其上多銅，其下多玉，無草木。滑水出焉，而西流注\circchar{于}諸\circchar{毗}之水，\annote{AN2.1}其中多滑魚，其\circchar{状}如鱓，赤背，\annote{AN2.2}其音如梧，\annote{AN2.3}食之已疣。\annote{AN2.4}其中多水馬，其状如馬，文臂牛尾，\annote{AN2.5}其音如呼。\annote{AN2.6}  {\color{black!70} \trans{Two hundred and fifty \emph{li} farther north lies Mount Qiuru (求如之山). Copper is abundant on its upper slopes and jade on its lower slopes, while no vegetation grows there. The Hua River (滑水) originates from this mountain and flows westward into the Zhubi River (諸毗之水). It abounds in \emph{hua} fish (滑魚), which resemble eels, have red backs, and make sounds like a person uttering "wu"; eating them cures warts. The river also contains many \emph{shuima} (水馬), which resemble horses, have patterned forelegs and ox tails, and neigh like a person calling out.}}

\colorbox{gray!20}{[AN2.5]:} 臂，前脚也。《周禮》曰：「馬黑脊而班臂，膢。」漢武元狩四年，燉煌渥洼水出焉，以為靈瑞者，即此𩔗也。  {\color{black!70}\trans{ \emph{Bi} (臂) refers to the forelegs. The \textit{Rites of Zhou} states: ``A horse with a black back and patterned forelegs is called \emph{lou}.'' In the fourth year of the Yuanshou era of Emperor Wu of Han, such creatures were said to have emerged from Wowa Lake in Dunhuang and were regarded as auspicious omens; they are the same kind of creature as the \emph{shuima}.}} \\
\hline
\colorbox{gray!20}{B-Text:} 又北二百五十里曰求如之山，$\cdots\cdots$滑水出焉，而西流注\circchar{于}諸\circchar{毗}之水。\annote{BN2.1}其中多滑魚，其\circchar{狀}如鱓，赤背，\annote{BN2.2}其音如梧，\annote{BN2.3}食之已疣。\annote{BN2.4}其中多水馬，其狀如馬，文臂牛尾，\annote{BN2.5}其音如呼。\annote{BN2.6}

\colorbox{gray!20}{[BN2.5]:} 郭曰：臂，前脚也。《周禮》曰$\cdots\cdots$ \boxlabel{任臣案}：漢馬出于余吾之水，……，皆水馬也……附記之《圖贊》曰：馬實龍精，爰出水類⋯⋯ {\color{black!70}\trans{Guo comments (郭曰): \emph{Bi} (臂) refers to the forelegs. The \textit{Rites of Zhou} states: ... \boxlabel{Ren Chen comments} (任臣案): According to the \textit{Book of Han}, horses also emerged from the waters of Yuwu ... they were all \emph{shuima} ... The \textit{Illustrated Eulogy} states: "The horse is in fact an essence of the dragon ..."}}  \\
\hline
\colorbox{gray!20}{C-Text:} 又北二百五十里曰求如之山，$\cdots\cdots$滑水出焉，而西流注\circchar{於}諸\circchar{𣬈}之水。\annote{CN2.1}其中多滑魚，其\circchar{狀}如鱓，赤背，其音如梧，食之已疣。\annote{CN2.2}其中多水馬，其狀如馬，文臂牛尾，其音如呼。\annote{CN2.3}

\colorbox{gray!20}{[CN2.3]:} 漢武帝元狩四年，得天馬於燉煌之渥洼水中。案：黑水經燉煌西流，此「滑」字與「渥洼」字音相近，殆即此水中也。 {\color{black!70}\trans{In the fourth year of the Yuanshou era of Emperor Wu of Han, a Heavenly Horse was obtained from Wowa Lake in Dunhuang. Comment: The Black River flows westward past Dunhuang. The pronunciation of hua (滑) is similar to that of Wowa (渥洼), so the Hua River mentioned here is probably the Black River.}} \\
\hline
\colorbox{gray!20}{D-Text:} 又北二百五十里曰求如之山。$\cdots\cdots$滑水\annote{DN2.1}出焉，而西流注\circchar{于}諸\circchar{𣬈}之水。\annote{DN2.2}其中多滑魚，\annote{DN2.3}其\circchar{狀}如鱓，赤背，\annote{DN2.4}其音如梧，\annote{DN2.5}食之已疣。\annote{DN2.6}其中多水馬，其狀如馬，文臂\annote{DN2.7}牛尾，\annote{DN2.8}其音如呼。\annote{DN2.9}

\colorbox{gray!20}{[DN2.8]:} 臂，前腳也。《周禮》曰$\cdots\cdots$  {\color{black!70}\trans{\emph{Bi} (臂) refers to the forelegs. The \textit{Rites of Zhou} states: … }}\\
\hline
\colorbox{gray!20}{E-Text:} 又北二百五十里曰求如之山，$\cdots\cdots$滑水\annote{EN2.1}出焉，而西流注\circchar{于}諸\circchar{𣬈}之水。\annote{EN2.2}其中多滑魚，\annote{EN2.3}其\circchar{狀}如鱓，赤背，\annote{EN2.4}其音如梧，\annote{EN2.5}食之已疣；\annote{EN2.6}其中多水馬，其狀如馬，文臂，牛尾，\annote{EN2.7}其音如呼。\annote{EN2.8}

\colorbox{gray!20}{[EN2.7]}: 臂，前脚也。《周禮》曰$\cdots\cdots$  \boxlabel{懿行案}：《内則》云：「馬黑脊而般臂，漏。」鄭注云：「漏當爲螻。如螻蛄臭也。」{\color{black!70}\trans{\emph{Bi} (臂) refers to the forelegs. The \textit{Rites of Zhou} states: … \boxlabel{Yi Xing comments} (懿行案): The \textit{Neize} states: "A horse with a black back and patterned forelegs is called lou (漏)." Zheng's note states: "\emph{Lou} (漏) should be written as \emph{lou} (螻), the odor of a mole cricket."}} \\
\hline
\end{tabularx}
\caption{Main text and commentaries for the ``Qiuru Mountain'' paragraph.}
\label{tab:qiuru_excerpt}
\end{table*}

\subsection{Commentary Hierarchy and Placement}
The main genres of ancient Chinese commentaries include \textit{zhuanzhu} (傳注), \textit{yishu} (義疏), and \textit{zhangju} (章句). Since the inline notes examined in this study belong to \textit{zhuanzhu} and \textit{yishu}, we briefly introduce these two genres below. \textit{Zhuanzhu} primarily focuses on lexical items. \textit{Yishu} is a form of sub-commentary whose explanatory target may lie in either the main text or an earlier commentary. They differ in commentary hierarchies. \textit{Zhuanzhu} contains a single commentary layer, while \textit{yishu} contains two commentary layers, whose second layer explains either the main text or the first layer. For clarity, we distinguish between \texttt{base-commentary} (b-commentary), inherited from earlier commentators (if present), and \texttt{core-commentary} (c-commentary), newly contributed layer in a given edition. Additionally, although all five editions comment on the same classic, their commentary placement differs owing to differences in exegetical conventions among commentators.

Among the five editions, Guo's Edition A is \textit{zhuanzhu} and the earliest extant commentary on the \textit{SHJ}, whereas the other four are \textit{yishu} that all draw on Guo's interpretations. The first layer of Edition B, D and E consists of Guo's annotations, separated from the second layer by formulaic expressions such as “任臣案”,“沅曰” and “懿行案” (“Renchen/Yuan/Yixing comments”). By contrast, Edition C integrates Guo's annotations with Wang's own interpretation into a single layer, due to the underdevelopment of evidential scholarship in Wang's time \citep{ChenShanHaiJingXueShuShiKaoLun2012}. Accordingly, Editions A and C contain only c-commentary, while the other three contain both.

The commentary placement in Edition C differs from the other four editions, placing commentaries at the end of complete sentences rather than after the associated main-text segments. This is evident from the placement of the commentary IDs in Table~\ref{tab:qiuru_excerpt}. Wang's placement strategy often cause one commentary to incorporate material from multiple entries in Guo's commentary.

\subsection{Contextual Exegesis}
The primary exegetical units of inline notes range from words and phrases to sentences and larger textual units \citep{WangXunGuXueYuanLi1996}. In our study, most commentaries target words, although their interpretations often extend beyond them because of contextual exegesis. The contextual dependency gives rise to two additional features that make identifying exegetical targets non-trivial and motivate the design of our framework.

First, different spans within a commentary may have different exegetical targets, both within and across commentary layers. For example, AN2.5 in Table~\ref{tab:qiuru_excerpt} first explains \emph{bi} (臂) and then shifts to the referent of \emph{shuima} (水馬). In EN2.7, the c-commentary explains \emph{lou} (膢), a term introduced in its b-commentary (derived from AN2.5) rather than the main text. This shows that interpreting c-commentaries may require reference to the hierarchical context formed by the main text and the preceding b-commentary. Similar phenomena also occur in \textit{Lunyu Jijie} \citep{HeLunYuLunYuJiJie} and the Baina edition of the \textit{Shiji} \citep{SimaShiJi}.

Second, the text segment to which a commentary is attached is not always its exegetical target. In AN2.5, the commentary is attached to “文臂牛尾”, yet its latter part interprets \emph{shuima}. Similarly, in \emph{xuncao} passage (Figure~\ref{fig:xuncao}), commentary DW9.7 is attached to “麻葉而方莖赤華而黑實” but ultimately targets \emph{xuncao}. Comparable examples also occur in the \textit{Zhuangzi} \citep{ZhuangZi}. We therefore define the attached text segment as the \texttt{anchor\_text}, distinguishing it from the actual exegetical target,

\subsection{Exegetical Functions}
Although inline notes must be interpreted in context, traditional Chinese philology also produced works such as the \textit{Erya} and \textit{Jingji Zuangu}, which reorganize exegetical knowledge extracted from diverse classical sources \citep{WangXunGuXueYuanLi1996}. Inspired by this tradition, we classify and, when necessary, decompose commentaries by their primary exegetical functions to facilitate downstream applications.

We propose a threefold classification: textual criticism (\texttt{TC}), lexical exegesis (\texttt{LE}), and referential exegesis (\texttt{RE}). \texttt{TC} addresses textual variants and textual emendation, \texttt{LE} explains the pronunciation, orthography, and meanings of words; and \texttt{RE} investigates named entities, including their attributes, identities, and historical transformations. Compared with earlier schemes \citep{HanXunGuXueZhuShiXue1989, HuangZhuShiLeiXing1999}, this framework is simpler and more applicable to downstream philological tasks. Notably, because referential exegesis historically developed from lexical exegesis, the boundary between the two is sometimes blurred. In this study, any commentary involving the external referent of a named entity is classified as \texttt{RE}. 

\subsection{Commentary Integration Criteria}
Although collected commentaries have been widely discussed in Chinese exegetics and philology \citep{HuangGuJiZhengLiGaiLun2001,GuoXunGuXue2005,YingStudyShijiJijie2005,ZhangDefinitionCollectedAnnotations2011,XuGuJiZhengLiShiLi2014,GuoZhongGuoGuDianWenXianXue}, little attention has been paid to the criteria for commentary integration. \citet{FengGuJiZhengLi2003} argues that commentaries collected under a \textit{jijie} entry should represent alternative interpretations of the same main-text segment, while \citet{MaReexaminationOriginTwoAnnotation2022} notes that integration also requires resolving differences in commentary placement as well as redundancy and contradiction across versions. The compilation of works such as \textit{Shiji} \citep{SimaShiJi} and \textit{Lunyu Jijie} \citep{HeLunYuLunYuJiJie} likewise suggest that commentaries are generally integrated by a shared textual segment. However, as discussed above, the contextual dependency of inline notes makes such segments difficult to identify. In practice, editorial convenience sometimes leads compilers to group commentaries on similar rather than identical segments. For example, \textit{Shanhaijing Jishi} places Wang's annotation on “南山經之首曰” together with Bi's annotation on “南山經之首” \citep{ZhouShanHaiJingJiShi2019}, sacrificing precision.

Operationally, our analysis of the \textit{SJ} commentaries and other representative \textit{jijie} works suggests that commentary integration is primarily determined by three factors: (1) shared main-text segments; (2) commentary placement; and (3) semantic similarity among commentaries. As inline notes usually follow the passages they explain, commentary placement is largely determined by the main text, although editorial conventions also affect placement (e.g., the different placement preferences between Guo and Wang). Semantic similarity is less informative for some commentaries because of their diverse content, but serves as a useful supplementary criterion for \textit{yishu} that quote or build upon earlier annotations.

\section{Computational Framework}
Based on the above analysis, our framework adopts shared main-text segments and semantic similarity between first-layer commentaries as the primary integration criteria, using commentary placement only during proofreading. Semantic similarity is included because Edition B-E are \textit{yishu} and have absorbed Edition A. Accordingly, the task is decomposed into two subtasks: (1) identifying the main-text segment associated with each commentary, and (2) grouping commentaries with the same main-text segment or highly similar first-layer commentary, based on semantic similarity. We address the first using LLMs and formulate the second as a cross-source mention clustering problem. To support downstream applications, we also identify each commentary's exegetical function in the first subtask.

\subsection{Two-step Prompt Chaining}
\begin{figure*}[t]
    \centering
    \includegraphics[width=\textwidth]{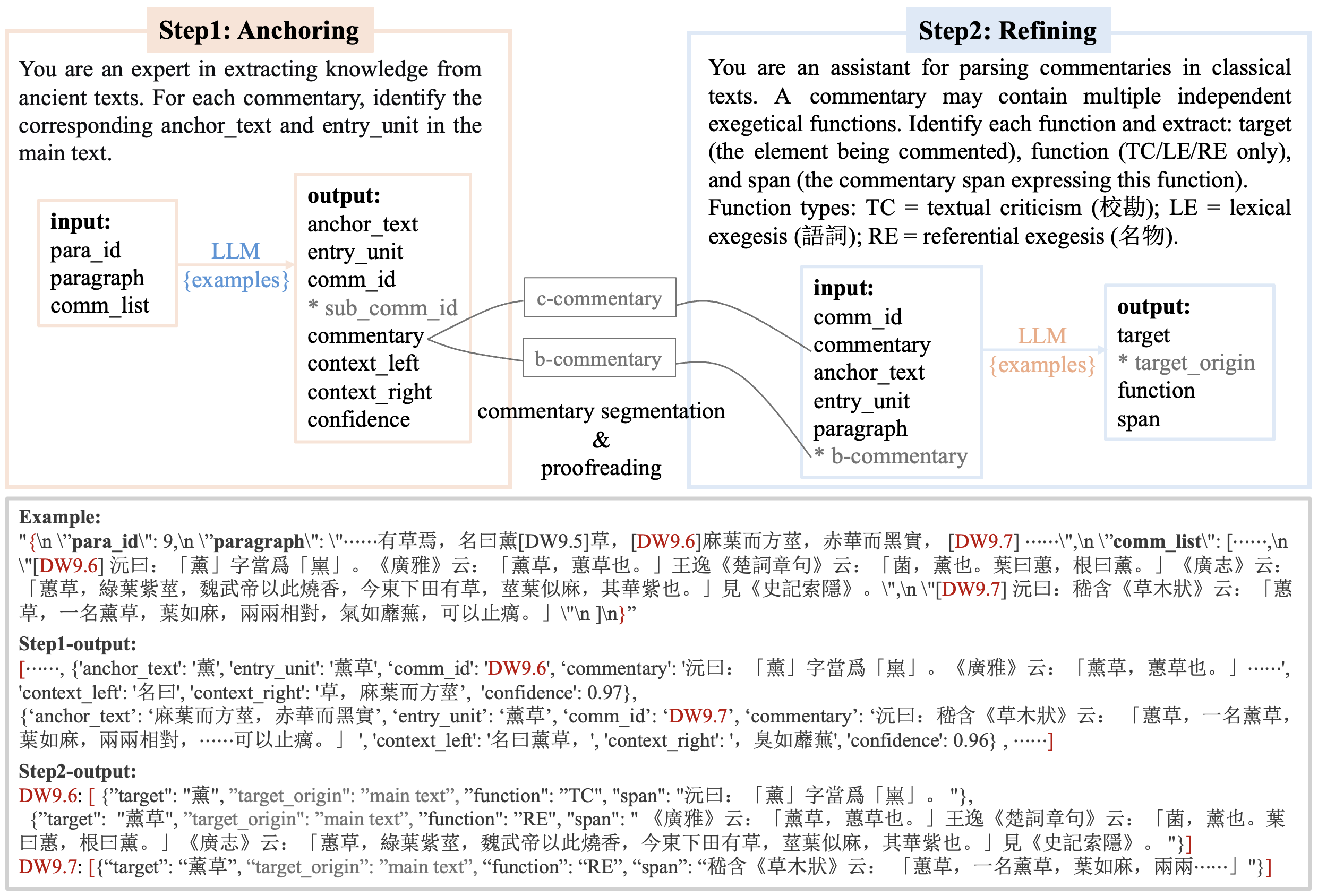}
    \caption{Two-step prompt chaining pipeline.}
    \label{fig:prompt}
\end{figure*}

Because a commentary may involve multiple exegetical targets and its \texttt{anchor\_text} is not always the ultimate target, we adopt a two-step prompt chaining strategy to identify the associated main-text segments. Prompt chaining has been shown to outperform a single prompt on complex tasks \citep{sunPromptChainingStepwise2024}. Figure~\ref{fig:prompt} illustrates the pipeline.

Each paragraph of each edition is processed independently. After paragraph IDs are assigned, the main text and commentaries are separated and converted into JSON as the pipeline input. Step 1 (Anchoring) identifies each commentary's \texttt{anchor\_text} and corresponding \texttt{entry\_unit} (primarily named entities). Step 2 (Refining) further determines the actual exegetical target and classifies or, when necessary, decomposes each commentary according to the threefold exegetical classification. For post hoc verification, auxiliary fields (e.g., \texttt{context\_left} and \texttt{context\_right}) are retained in the output. Commentaries with two layers are separated after Step 1 so that Step 2 operates only on the c-commentary. The example at the bottom of Figure~\ref{fig:prompt} illustrates the two challenges addressed by the framework: multiple exegetical targets within a single commentary (DW9.6) and the mismatch between the \texttt{anchor\_text} and the actual exegetical target (DW9.7).

Because the five editions differ in commentary hierarchy and placement, we designed two prompt variants (a and b). Variant b adds additional fields (marked with asterisks in Figure~\ref{fig:prompt}) for edition-specific features, including commentary splitting and b-commentary-dependent interpretation.

\subsection{Cross-source Mention Clustering}
Clustering is based on the semantic similarity of either the associated main-text segment or the first-layer commentary. The associated main-text segment may be the \texttt{anchor\_text}, \texttt{entry\_unit} or target identified in Subtask 1. Partly due to the high density of named entities, commentaries like DW9.7 that are attached to one segment but target another are very common in \textit{SJ}. We therefore use the \texttt{anchor\_text} for clustering. Using the \texttt{entry\_unit} or target alone may merge commentaries with different exegetical focuses or decomposed commentaries.

Algorithm~\ref{alg:trcuf} presents the cross-source mention clustering procedure. Commentaries from the same \textit{SJ} paragraph across the five editions are processed as a single batch. For each commentary, the top-k candidates are retrieved according to the semantic similarity of \texttt{anchor\_text} contextual embeddings or first-layer commentary embeddings. Clustering is then performed using a Union-Find algorithm with three constraints: (1) cluster-size limits, (2) complete linkage for commentary-similarity merging, and (3) source uniqueness. Similar strategies have been adopted in record linkage \citep{guoRecordLinkageUniqueness2010}, entity resolution \citep{bhattacharyaCollectiveEntityResolution2007, kardesGraphbasedApproachesOrganization2013}, and mention clustering \citep{zhangNeuralCoreferenceResolution2018}. In our framework, these constraints primarily prevent bridging effects.

\begin{algorithm}[t]
\small
\caption{Two-Route Union-Find Clustering.}
\label{alg:trcuf}
\begin{algorithmic}[1]
\STATE Initialize each row as a singleton cluster.

\STATE \textbf{Phase 1: Commentary Merge}
\FOR{each candidate pair $(i,j)$}
    \STATE $C_i \leftarrow \mathrm{cluster}(i)$, $C_j \leftarrow \mathrm{cluster}(j)$
    \IF{$C_i \neq C_j$ and $\mathrm{comm\_sim}(i,j) \geq T_c$}
        \IF{$\forall x \in C_i, \forall y \in C_j: \mathrm{comm\_sim}(x,y) \geq T_c$ and $\mathrm{source}(C_i) \cap \mathrm{source}(C_j) = \emptyset$}
            \STATE merge$(C_i, C_j)$
        \ENDIF
    \ENDIF
\ENDFOR

\STATE \textbf{Phase 2: Anchor-text Merge}
\FOR{each candidate pair $(i,j)$}
    \STATE $C_i \leftarrow \mathrm{cluster}(i)$, $C_j \leftarrow \mathrm{cluster}(j)$
    \IF{$C_i \neq C_j$ and $\mathrm{anchor\_sim}(i,j) \geq T_t$}
        \IF{$(\mathrm{source}(C_i) \cap \mathrm{source}(C_j)) \subseteq \{C\}$}
            \STATE merge$(C_i, C_j)$
        \ENDIF
    \ENDIF
\ENDFOR

\STATE Assign each row the ID of its final cluster.
\STATE Output clustered rows with \texttt{group\_id}.
\end{algorithmic}
\end{algorithm}

\section{Experiments}
Since all five editions annotate the \textit{SJ}, we compare them at multiple levels using character counts and character-frequency distributions to characterize their similarity and inform experiments. The main texts and Guo's commentaries exhibit over 96\% similarity in character-frequency distributions. Similarity decreases as later commentaries are incorporated, but remains above 75\%. This high similarity makes model selection and clustering thresholds particularly important. 

\subsection{Model Selection}
We evaluate three LLMs\footnote{DeepSeek-Chat, Kimi-K2-0905-Preview, and GPT-5-mini.} for Subtask 1 and seven Classical Chinese BERT models \footnote{SikuBERT, SikuRoBERTa \citep{WangConstructionApplicationPretrained2022}, BERT-Ancient-Chinese \citep{wangUncertaintybasedRetrievalFramework2022}, RoBERTa-Classical-Chinese-Base-Char \citep{YasuokaRobertaclassicalchinesebase}, RoBERTa-Classical-Chinese-Large-Char \citep{YasuokaRobertaclassicalchineselarge}, GuwenBERT-Base \citep{EthanGuwenbertbase}, GuwenBERT-Large \citep{EthanGuwenbertlarge}.} for Subtask 2 on manually curated datasets. Based on their performance, we select DeepSeek and SikuRoBERTa, respectively. For SikuRoBERTa, we average the last five hidden layers for \texttt{anchor\_text} contextual embeddings, the last two for commentary embeddings, and use the corresponding \texttt{pos\_p05} and \texttt{pos\_mean} values as clustering thresholds.

\subsection{Evaluation}
The extraction and clustering results for the \textit{Classic of Northern Mountains} section (18.044\% of the dataset) were manually verified and used as the gold standard for evaluation. For Subtask 1, we report the accuracy of each step , their weighted average (E-E Acc.), and the mean Jaccard score for Step 2, which evaluates the identification of multiple exegetical functions within a single commentary. Table~\ref{tab:subtask1} summarizes the results. The overall accuracy is 84\%, with errors mainly caused by the inaccurate \texttt{anchor\_text} extraction and incorrect decomposition of exegetical functions.

\begin{table}[t]
\centering
\small
\begin{tabular}{cccc}
\hline
S1 Acc. & S2 Acc. & S2 Jacc. & E2E Acc. \\
\hline
0.827 & 0.853 & 0.848 & 0.840 \\
\hline
\end{tabular}
\caption{Extraction performance of Subtask 1.}
\label{tab:subtask1}
\end{table}

We evaluate clustering performance using standard coreference metrics: MUC \citep{vilainModelTheoreticCoreferenceScoring1995}, $B^3$ \citep{baggaAlgorithmsScoringCoreference1998}, and CEAF$_{\phi_4}$ \citep{luoCoreferenceResolutionPerformance2005}, and report the CoNLL score \citep{pradhanCoNLL2011SharedTask2011, pradhanCoNLL2012SharedTask2012} as the their average F$_1$. Pairwise scores are reported to evaluate pairwise linking. Table~\ref{tab:subtask2} shows that precision and recall exceed 95\% across all metrics, and the CoNLL F$_1$ exceeds 97\%. Errors are mainly caused by the textual variants in \texttt{ancho\_text} and Wang's commentaries.

\begin{table}[t]
\centering
\small
\begin{tabular}{lcccc}
\hline
Metric & Pairwise & $B^3$ & MUC & CEAF$_{\phi_4}$ \\
\hline
Precision & 0.993 & 0.995 & 0.993 & 0.972 \\
Recall    & 0.953 & 0.967 & 0.963 & 0.972 \\
F$_1$     & 0.972 & 0.981 & 0.978 & 0.972 \\
\hline
CoNLL F$_1$ & \multicolumn{4}{c}{0.977} \\
\hline
\end{tabular}
\caption{Clustering performance of Subtask 2.}
\label{tab:subtask2}
\end{table}

\subsection{Results}
Table~\ref{tab:statistics} summarizes the extraction and clustering statistics. More than 78\% of \textit{SJ} paragraphs contain commentaries. The numbers of \texttt{Comm.} and \texttt{C-comm.} reflect the accumulation of exegetical knowledge over time, while the distributions of \texttt{TC}, \texttt{LE}, and \texttt{RE} further reveal a shift in exegetical practice. Guo's commentary is dominated by \texttt{LE}, whereas Qing-dynasty commentaries contain fewer pure \texttt{LE} cases and many more \texttt{RE} or \texttt{TC}, likely refelcting the rise of evidential scholarship, in which \texttt{LE} increasingly served \texttt{RE} and \texttt{TC} rather than functioning independently. The \texttt{Split} rate indicates that each c-commentary contains an average of 1.16-1.26 exegetical functions. Finally, the c-commentaries (5,463 in total) are grouped into 2,440 clusters, averaging 2.24 commentaries per semantic anchor.

\begin{table}[t]
\centering
\small
\begin{tabular}{lccccc}
\hline
Metric & A & B & C & D & E \\
\hline
Ann. (\%) & 78.3 & 85.6 & 81.0 & 86.0 & 90.4 \\
Comm.       & 974  & 1374 & 843  & 1517 & 1686 \\
C-comm.     & 974  & 962  & 1064 & 970  & 1493 \\
TC          & 146  & 172  & 135  & 349  & 699 \\
LE          & 529  & 85   & 496  & 149  & 206 \\
RE          & 556  & 859  & 712  & 628  & 838 \\
Split       & 1.264 & 1.160 & 1.262 & 1.161 & 1.167 \\
\hline
Clusters & \multicolumn{5}{c}{2440} \\
\hline
\end{tabular}
\caption{Extraction and clustering statistics.}
\label{tab:statistics}
%\vspace{1mm}
\begin{minipage}{\linewidth}
\footnotesize
\emph{Note:} \texttt{Ann.} denotes the percentage of \textit{SJ} paragraphs with commentaries. \texttt{Comm.} and \texttt{C-comm.} denote the numbers of commentaries and c-commentaries. For Editions A and C, they would theoretically be identical, but \texttt{C-comm.} is larger in Edition C because some commentaries are split during Step 1 of prompt chaining. \texttt{TC}, \texttt{LE}, and \texttt{RE} report the distribution of the three exegetical functions among c-commentaries. \texttt{Split} rate = (\texttt{TC} + \texttt{LE} + \texttt{RE}) / \texttt{C-comm.}, indicating the average diversity of exegetical functions per c-commentary.
\end{minipage}
\vspace{-3mm}
\end{table}

\section{Implications}
\begin{figure*}[t]
    \centering
    \includegraphics[width=\textwidth]{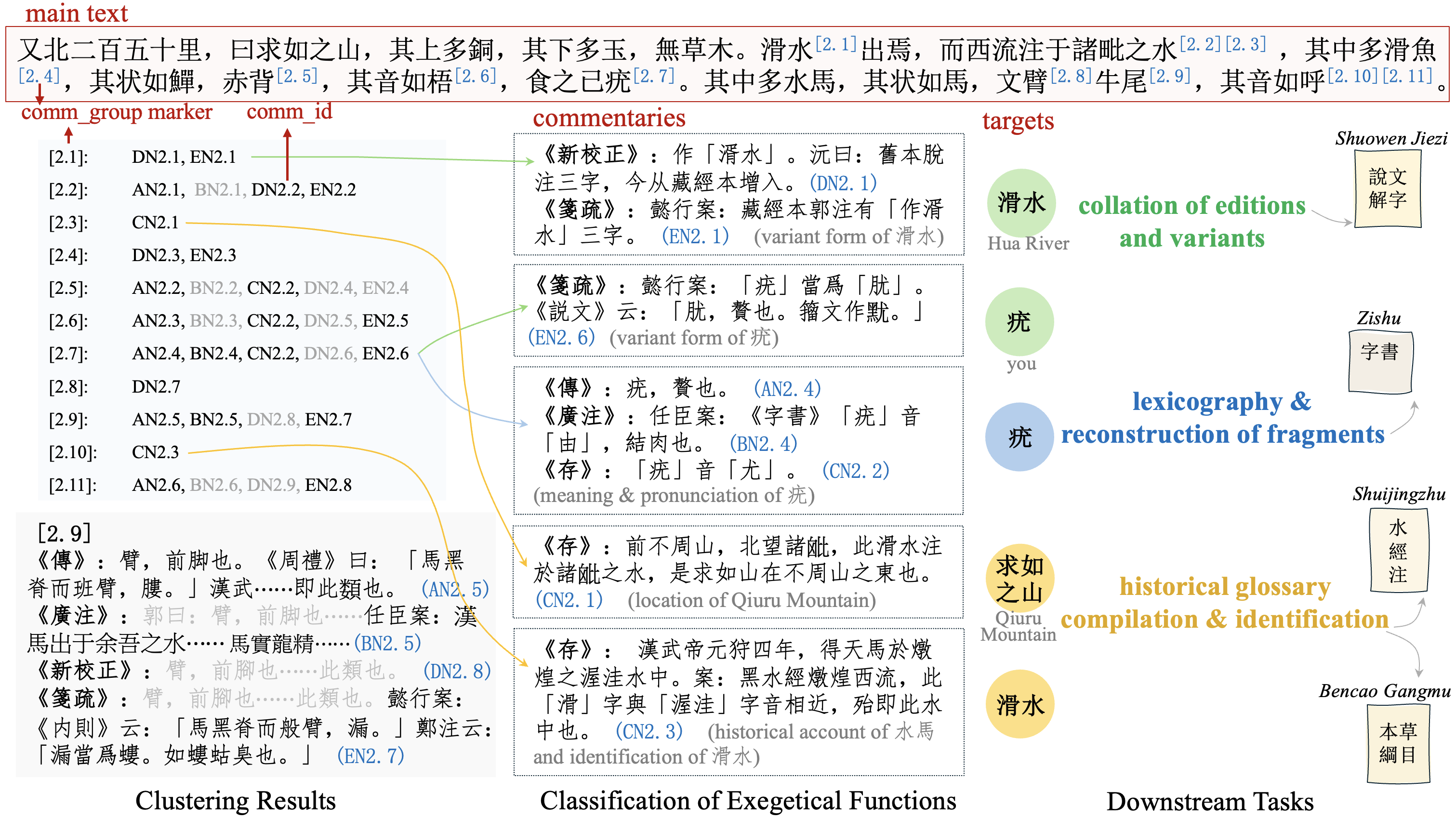}
    \caption{Experimental results from the “Mountain Qiuru” paragraph and their possible applications.}
    \label{fig:implications}
\end{figure*}

The implications of this study extend beyond the \textit{SJ}. Figure~\ref{fig:implications} illustrates the experimental results for the “Qiuru Mountain” paragraph in Table~\ref{tab:qiuru_excerpt}. The left panel shows commentary integration results \footnote{Commentaries consisting solely of Guo's commentary (without a c-commentary) or repeated occurrences of Guo's commentary are marked in light gray.}, the middle panel the threefold classification of exegetical functions \footnote{Rectangles represent commentaries and circles represent their corresponding exegetical targets; green, blue, and yellow denote TC, LE, and RE, respectively.}, and the right panel potential philological downstream applications. In the main text above, commentary-group markers (\texttt{comm\_group marker}) indicate both commentary placement and clustering assignments. Using this example, we discuss the broader implications of our framework from four perspectives: context-preserving commentary integration, exegetical function classification for philological downstream tasks, edition genealogy through variant character analysis, and potential NLP applications.

\subsection{Context-Preserving Integration}
Our framework groups commentaries that share the same \texttt{anchor\_text} or have similar first-layer annotations and orders them chronologically by edition. Even after duplicate b-commentaries are removed, the interpretive context formed by the \texttt{main text - b-commentary - c-commentary} hierarchy is preserved. For example, all commentaries in group [2.9] of Figure~\ref{fig:implications} were originally attached to the same segment (see Table~\ref{tab:qiuru_excerpt}), and the c-commentary of EN2.7 remains fully interpretable through AN2.5. Cases that do not satisfy these criteria are resolved through manual philological verification.

\subsection{Philological Downstream Tasks}
The proposed threefold classification of exegetical functions, combined with data linking, supports downstream tasks such as textual collation and variant collection, lexicography and lost-text reconstruction, and historical encyclopedia compilation and historical entity identification.

\texttt{TC} commentaries support textual collation and variant collection by recording textual variants and editorial corrections. For example, DN2.1 and EN2.1 identify \emph{huashui} (滑水) as a variant of \emph{xushui} (湑水), while EN2.6 relates 疣 to 肬 in the \textit{Shuowen Jiezi}. Such information is essential for reconstructing reliable texts, as demonstrated in recent studies on the \textit{Shuijingzhu} \citep{LiShuiJingZhuBanBen2021}.

\texttt{LE} commentaries support lexicography and lost-text reconstruction by explaining word meanings and pronunciations while preserving quotations from other philological works. For example, AN2.4 and CN2.2 explain the meaning and pronunciation of 疣, offering evidence for the diachronic development of Classical Chinese vocabulary. BN2.4 cites the lost \textit{Zishu}. Collecting such quotations at scale facilitates the reconstruction of this work, as illustrated by \citet{WangZishu2016}. Similar quotation-based approaches have also been used to medieval fragments \citep{bertiLinkedFragmentTEI2014/Dec/28}.

\texttt{RE} commentaries support historical encyclopedia compilation and historical entity identification by preserving investigations of historical entities and institutions. Examples include CN2.1 on the relationship between \emph{Qiuru Mountain} and \emph{Buzhou Mountain}, and CN2.3 on the identity of \emph{shuima} and the geography of \emph{huashui}. Such evidence can be integrated with other classical sources for large-scale historical knowledge reconstruction. Comparable uses of commentarial evidence have been reported for named-entity disambiguation in the \textit{Zuozhuan} \citep{LiDigitalHumanityBased2020} and LLM-based geocoding of Early China toponyms \citep{chenGeocodingWorldUnearthing2026}.

\subsection{Edition Genealogy Analysis}
\begin{figure}[t]
    \centering
    \includegraphics[width=\columnwidth]{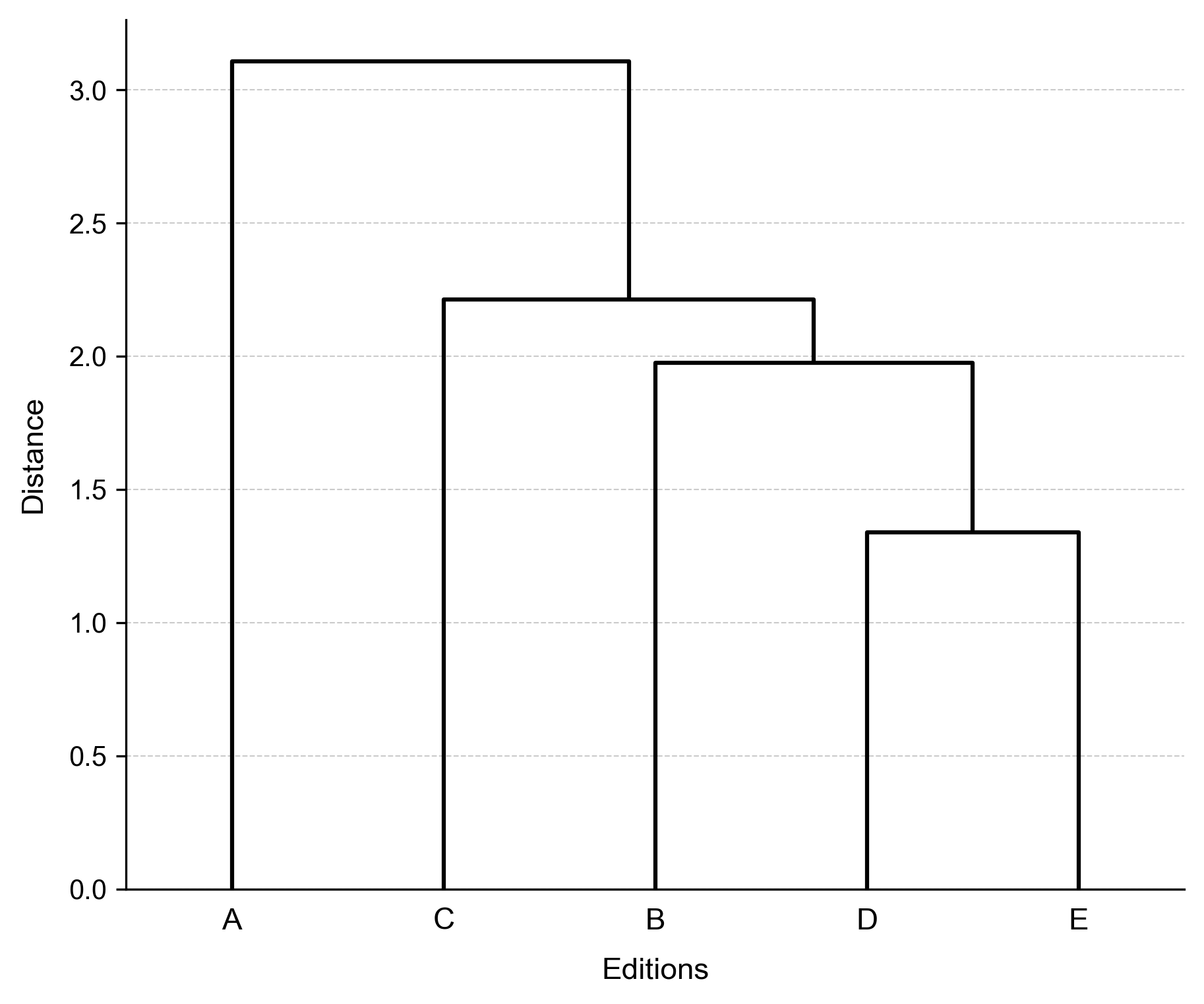}
    \caption{Edition clustering by variant pairs.}
    \label{fig:genealogy}
\end{figure}

The organization of traditional commentaries often requires careful comparison across textual editions and commentary layers \citep{NiXiaoKanXue2004}. As Table~\ref{tab:qiuru_excerpt} shows, the five editions exhibit a number of character variants in the main text. Based on the character frequency analysis in Section 5, we further cluster the five editions using the 20 most frequent variant-character pairs as features. The results suggest that variant-character patterns can serve as a quantitative indicator of edition genealogy. The resulting dendrogram in Figure~\ref{fig:genealogy} closely agrees with the philological analysis of \citet{ZhangShanHaiJingYanJiu2007}: Editions D and E derive from the Ming-dynasty woodblock edition collated by Xiangyin (項絪), Editions B and C represent two other branches of the Ming woodblock tradition, and Edition A was reproduced from a Ming imperial edition.

\subsection{Potential NLP Applications}
As discussed above, the organized exegetical knowledge produced by our framework can provide foundational knowledge for many NLP tasks, such as CWS and NER. It also enables the adaptation of other NLP tasks, such as text summarization and textual entailment, to collected commentaries, supporting applications including commentary summarization and cross-commentator consistency analysis.

\section{Conclusion}
This paper formulates collected commentary compilation as an NLP task and proposes a computational framework for preserving the contextual dependency of inline notes and organizing historical exegetical knowledge. The main limitation of this work is that our experiments are limited to the \textit{SJ}, although the proposed framework is generalizable to other ancient texts. Future work includes extending the framework to a broader range of ancient texts, investigating its contribution to related NLP tasks, and reconstructing the natural-historical landscape of the \textit{SJ} based on the resulting structured knowledge.

%\bibliography{tacl2021}
\bibliographystyle{acl_natbib}
\bibliography{references}

\end{document}